\pdfoutput=1

\documentclass[11pt]{article}

\usepackage[final]{acl}

\usepackage{times}
\usepackage{amsmath}
\usepackage{latexsym}
\usepackage{todonotes}
\usepackage{hyperref}
\usepackage{cleveref}

\usepackage[T1]{fontenc}

\usepackage[utf8]{inputenc}

\usepackage{microtype}

\usepackage{inconsolata}

\usepackage{graphicx}

\usepackage{adjustbox}
\usepackage{amsfonts}
\usepackage{colortbl}
\usepackage{xcolor}
\usepackage{booktabs}
\usepackage{makecell}
\usepackage{comment}

\title{Optimizing LLMs for Italian: Reducing Token Fertility and Enhancing Efficiency Through Vocabulary Adaptation}

\author{
    \textbf{Luca Moroni}\textsuperscript{1}\thanks{\ Those authors contributed equally.}, \textbf{Giovanni Puccetti}\textsuperscript{2}$^*$, \textbf{Pere-Lluis Huguet Cabot}\textsuperscript{1}, \textbf{Andrei Stefan Bejgu}\textsuperscript{4}
    \\
    \textbf{Edoardo Barba}\textsuperscript{1}, \textbf{Alessio Miaschi}\textsuperscript{3}\\
    \textbf{Felice Dell'Orletta}\textsuperscript{3}, \textbf{Andrea Esuli}\textsuperscript{2}, \textbf{Roberto Navigli}\textsuperscript{1}
    \\
    \textsuperscript{1}Sapienza University of Rome\;
    \texttt{\{surname\}@diag.uniroma1.it}\\
    \textsuperscript{2}ISTI-CNR\;
    \texttt{\{name.surname\}@isti.cnr.it}\\
    \textsuperscript{3}ILC-CNR\;
    \texttt{\{name.surname\}@ilc.cnr.it}\\
    \textsuperscript{4}Babelscape\;
    \texttt{\{surname\}@babelscape.com}
}

\newcommand{\mistral}{\emph{Mistral-7B-v0.1}}
\newcommand{\llama}{\emph{Llama-3.1-8B}}
\newcommand{\minervas}{\textit{Minerva-350M}}
\newcommand{\minervam}{\textit{Minerva-1B}}
\newcommand{\minerval}{\textit{Minerva-3B}}
\newcommand{\minervallms}{\textit{Minerva-LLMs}}
\newcommand{\llamantino}{\textit{LLaMAntino-2-LLMs}}
\newcommand{\anita}{\textit{LLaMAntino-3-ANITA-8B-Inst-DPO-ITA}}

\begin{document}

\definecolor{low}{rgb}{0.8, 0.9, 1} %
\definecolor{mid}{rgb}{1, 1, 0.8}   %
\definecolor{high}{rgb}{1, 0.8, 0.8} %

\maketitle
\begin{abstract}
The number of pretrained Large Language Models (LLMs) is increasing steadily, though the majority are designed predominantly for the English language. 
While state-of-the-art LLMs can handle other languages, due to language contamination or some degree of multilingual pretraining data, they are not optimized for non-English languages, leading to inefficient encoding (high token "fertility") and slower inference speed.
In this work, we thoroughly compare a variety of vocabulary adaptation techniques for optimizing English LLMs for the Italian language, and put forward Semantic Alignment Vocabulary Adaptation (SAVA), a novel method that leverages neural mapping for vocabulary substitution. SAVA achieves competitive performance across multiple downstream tasks, enhancing grounded alignment strategies. We adapt two LLMs: \mistral{}, reducing token fertility by 25\%, and \llama{}, optimizing the vocabulary and reducing the number of parameters by 1 billion. We show that, following the adaptation of the vocabulary, these models can recover their performance with a relatively limited stage of continual training on the target language. Finally, we test the capabilities of the adapted models on various multi-choice and generative tasks.\footnote{We release our code and models at \url{https://github.com/SapienzaNLP/sava}}
\end{abstract}

\section{Introduction}

\begin{figure}[t]
\centering
  \includegraphics[width=1\columnwidth, trim=0 0 0 0, clip]{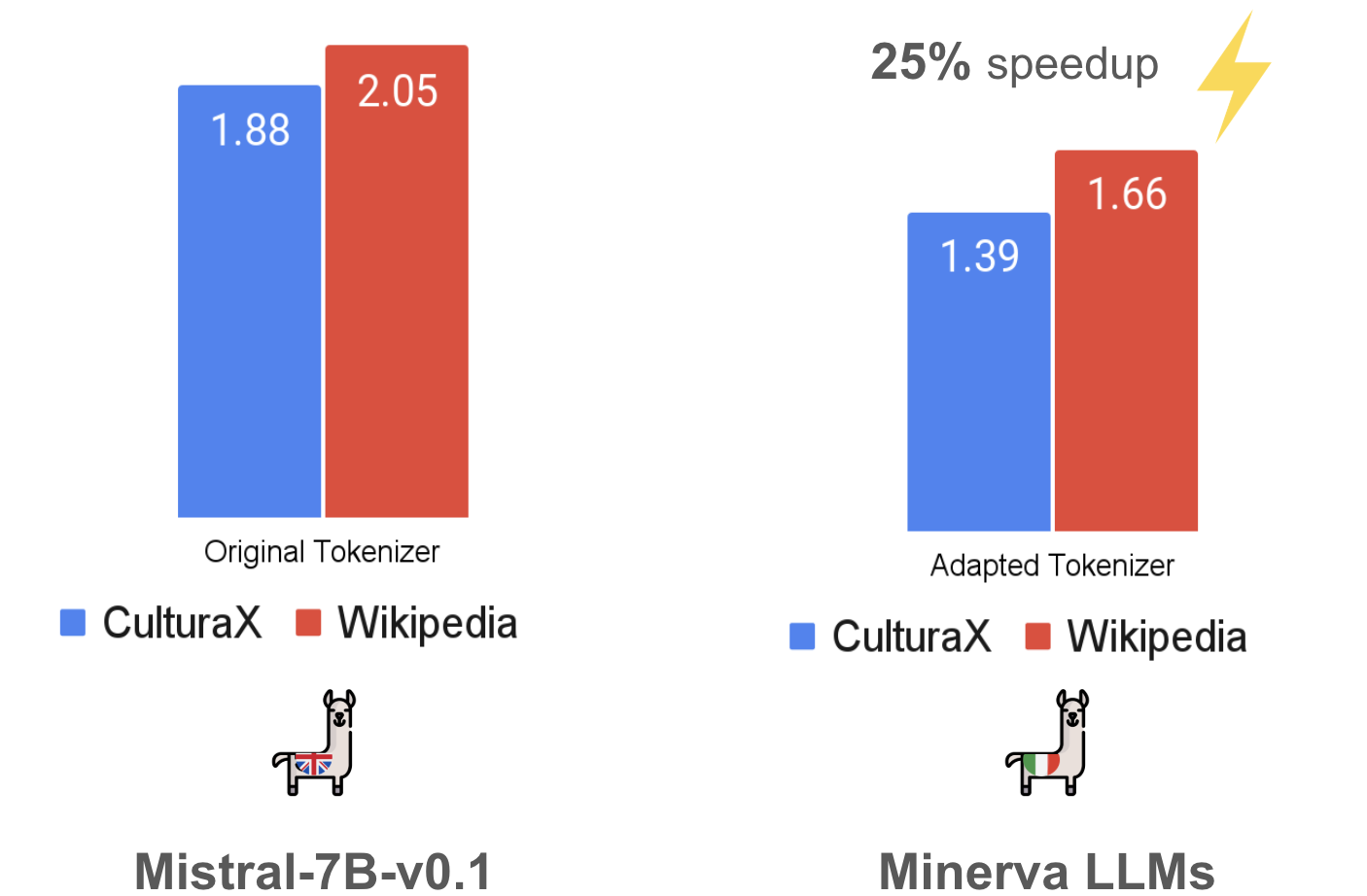}
  \caption{Fertility for two different tokenizers, \textit{Mistral-7B-v0.1} (left) and \textit{Minerva} (right), over Italian texts from  CulturaX (blue) and Wikipedia (red).}

  \label{fig:mistral_italian_graphical}
\end{figure}

Large Language Models (LLMs) have gained immense popularity and are increasingly being utilized across a wide range of applications \citep{radford2019language,kojima2022large}.
Despite their impressive performance, these models are mainly English-centric, that is, most state-of-the-art models are designed and pre-trained on datasets with a primary focus on English \citep{jiang2023mistral,dubey2024llama,team2024gemma}.
Although native multilingual models -- i.e. fully pre-trained in multiple target languages -- have been released over the years \citep{le2023bloom}, they still fall short of achieving performance levels comparable to models pre-trained in English.
The primary challenge is addressing underrepresented languages, where large, clean, open-access text corpora are often scarce \citep{together2023redpajama,nguyen-etal-2024-culturax}. This scarcity is problematic because models require vast amounts of high-quality data to achieve satisfactory performance \citep{hoffmann2022training}. Moreover, multilingual models generally reach sub-optimal performance due to the well-known \textit{curse of multilinguality} \citep{conneau-etal-2020-unsupervised}.

A promising solution to these challenges is the adaptation of pretrained English LLMs to other languages \citep{chau-etal-2020-parsing}. Recent studies highlight that fine-tuning English-centric models to support other languages yields substantial benefits, allowing for efficient adaptation while minimizing computational resources and training time. This method reduces both the training budget and the number of tokens required, demonstrating competitive performance even in low-resource scenarios \citep{koto-etal-2021-indobertweet,minixhofer-etal-2022-wechsel,gee-etal-2022-fast, ostendorff2023efficient}.

Another important aspect, alongside the downstream performance of language models, is the \textit{tokenizer’s fertility} in target languages. LLMs rely on a tokenizer, which is trained on a mix of text (either LLM's training data or not), transforming raw text into word-piece tokens; \emph{fertility} is the average number of tokens in which a word is split \citep{brown-etal-1993-mathematics}. The fertility of a tokenizer is highly sensitive to the language and the type of text it was trained on, as well as the text on which the fertility itself is measured. \Cref{fig:mistral_italian_graphical} shows an example of this phenomenon comparing the fertility of \minervallms{}, a family of Italian-first LLMs \citep{orlando2024minerva}, and \mistral{}, an English-first LLM \citep{jiang2023mistral}, on two corpora in Italian.  

In this work, we explore the adaptation of two state-of-the-art English LLMs to the Italian language using both vocabulary adaptation and continual learning. Additionally, we introduce a novel vocabulary adaptation technique called \textit{Semantic Alignment Vocabulary Adaptation (SAVA)} and conduct a comprehensive comparison with recent approaches \citep{gee-etal-2022-fast, ostendorff2023efficient}, examining the impact of vocabulary substitution on model performance throughout the adaptation process.
After the adaptation of the vocabulary,  when tokenizing Italian texts, we are able to reduce the fertility of \mistral{} by 25\% and of \llama{} by 16\%. 
As regards \mistral{}, we do not increase its vocabulary size or model parameters, while for \llama{}, we effectively reduce its vocabulary size by 75\% thereby reducing the final model size by 10\%.
Overall, we reduce memory and compute footprint of the models. To summarize, the main contributions are:
\begin{itemize}
  \item Introducing an effective approach for adapting tokenizers and vocabularies of generative models, leading to competitive performance over existing methods across several downstream benchmarks;
  \item Providing a detailed comparative analysis of various tokenizer adaptation techniques, with a focus on continual training in low- to mid-resource scenarios.
  \item   Analyzing the embedding representations learned through different adaptation techniques, offering a deeper understanding of how vocabulary modifications impact model performance and generalization.
\end{itemize}

\section{Related Work}

\paragraph{Language-Adaptive Pretraining} Designing LLMs in a target language and thus training them from scratch is the best approach to obtain an adequate token fertility from the outset and minimize interference from pretrained data on different languages. 
However, this approach is often impractical, especially in low-resource settings and on a low computational budget. For this reason, several recent studies \citep{de-vries-nissim-2021-good,gee-etal-2022-fast,csaki-etal-2024-sambalingo} have focused on the adaptation of pretrained LLMs to new languages. Pretrained LLMs can be adapted to a specific language using a small quantity data compared to what is needed in the pretraining stage. A straightforward approach to achieving this is Language-Adaptive Pre-Training (LAPT), utilized by \citet{chau-etal-2020-parsing} in a multilingual setting where they tested continual training of multilingual LLMs on target languages. Interestingly, LAPT was previously proposed on encoder-only architectures by \citet{gururangan-etal-2020-dont}, where they successfully adapted RoBERTa \citep{zhuang-etal-2021-robustly} models in a biomedical domain. In LAPT, models do not undergo any structural change to their architecture.
This usually results in performance improvements, however it does not address the limitations of using a sub-optimal tokenizer that is less suited to the encoding of different languages. 
Regarding LAPT research in English-to-Italian models, there have been several attempts, most notably \llamantino{}, which is a fine-tuning of LLama 2 on Italian translated conversations \citep{basile2023llamantino}, and \anita{}, a more recent effort that is built upon \textit{Llama-3-8B} using a similar approach \citep{polignano2024advanced}.

\paragraph{Vocabulary Adaptation Techniques} To tackle the fertility issue, recent research has focused on improving language adaptation by modifying the tokenizer and vocabulary of pretrained LLMs to better fit the target language. Several efforts in this area have shown the effectiveness 
of vocabulary adaptation techniques. \citet{minixhofer-etal-2022-wechsel} and \citet{liu-etal-2024-ofa} propose to replace the tokenizer of a pretrained LLM, along with its corresponding embedding layer, relying on a bilingual dictionary-based, or graph-based, token mapping. 
Generally, the main difference between various vocabulary adaptation techniques lies in how the embedding space of the respective model is initialized during adaptation. More effort was made by \citet{ostendorff2023efficient,dobler-de-melo-2023-focus} who use the embeddings from a helper model trained alongside the desired tokenizer. They utilize geometrical similarities in the helper model's embedding structure to initialize the tokens' representations of the target model effectively.
In parallel, \citet{gee-etal-2022-fast} proposed a simple heuristic, initializing target vocabulary tokens as the average of their corresponding sub-tokens in the source vocabulary. Another study by \citet{koto-etal-2021-indobertweet} put forward an adaptation technique, they rely on \texttt{FastText}\footnote{\url{https://fasttext.cc/}} embedding space to learn a linear mapping, to perform vocabulary adaptation of BERT-based models.

Unlike previous studies, we thoroughly analyze existing adaptation heuristics, focusing on decoder-only generative models adapted to Italian. We present a novel heuristic that utilizes a helper embedding space, optimized for the target language, to map and initialize target vocabulary tokens.

\section{Methodology}
In this section, we formalize the methodologies used to adapt pretrained LLMs to a target language. The following subsections outline the techniques employed to modify the vocabulary of pretrained LLMs and describe the process of adapting them to a target language. Finally, we describe the last step of the adaptation, that is, the continual training step.

\subsection{Vocabulary Adaptation}

All the vocabulary adaptation methodologies share a similar objective: substituting the tokenizer and its vocabulary, and replacing the model embeddings (both embedding module and language model head) with one more suited for the target language.

In our setting, we have a source pretrained LLM, $M_s$, with its embedding matrix $E_s$\footnote{Here, we assume tied-weights, i.e., shared embedding module and language model head. When this is not the case, the approach is symmetric, as if there were two embedding matrices.}, tokenizer $T_s$, and vocabulary $V_s$. To adapt our model to a target language, we have a target tokenizer $T_t$ and vocabulary $V_t$ suited to encoding texts in the target language, which we want to make $M_s$ compatible with. In some cases, we also have access to a helper model, $M_h$, which is an LLM, usually smaller than $M_s$, whose embeddings are noted with $E_h$. The helper model is trained using $T_t$ and $V_t$. We use the superscript notation $E^{t_i}$ to indicate the representation of the token $t_i$ on the matrix embedding $E$.

The objective is to adapt the source model embeddings $E_s$, which all these methods operate on, so as to obtain $E_t$ based on the target tokenizer $T_t$ and the target vocabulary $V_t$.

First, the target embeddings are initialized by keeping the same representation from $E_s$ for the tokens in the intersection of both vocabularies, while a function $g$ is applied to the remaining ones:
\begin{equation*}
    E_t^{t_i} = \begin{cases}
        g(t_i, \cdot{}), & t_i \in V_t \setminus V_s \\
        E_s^{t_i}, & t_i \in V_s \cap V_t
    \end{cases}
\end{equation*}
The difference between these methods lies in $g$, used to initialize the tokens that are in $V_t$ and not in $V_s$. This function has access to the source embeddings $E_s$, vocabulary $V_s$ and tokenizer $T_s$ and possibly the embeddings, vocabulary, and tokenizer of a helper model $E_h$, $V_h$ and $T_h$, respectively. 

Therefore, each method is defined by its respective $g$ function, as detailed below.

\paragraph{Random} As a baseline approach, we initialize the tokens outside the intersection with a random representation, given by a normal distribution, with the mean and the variance defined by the source embedding space:
$$g_{random}(t_i, E_s) = \mathcal{N}(\mu(E_s), \sigma^2(E_s))$$

\paragraph{FVT} \citet{gee-etal-2022-fast} introduced Fast Vocabulary Transfer (FVT) for vocabulary adaptation, which consists of an efficient way to initialize the intersection tokens in the target embedding space. Here, each target token is computed with the average of the embedding source tokens given by the source tokenizer, i.e. the resulting tokens when we tokenize the target token $t_i$ with $T_s$:
$$g_{fvt}(t_i, E_s, T_s) = \frac{1}{\lvert T_s(t_i)\rvert} \cdot \sum_{t_j \in T_s(t_i)} E_s^{t_j}.$$

\paragraph{CLP} \citet{ostendorff2023efficient} and in parallel \citet{dobler-de-melo-2023-focus} introduced a heuristic to initialize out-of-inventory tokens relying on the space structure of the helper embedding space. Both approaches compute similarity scores between the tokens in $V_t \setminus V_s$ against the ones in $V_t \cap V_s$, on the embedding space of the helper model $E_h$. Such similarities are used to construct a representation of the out-of-inventory tokens in the target embedding matrix $E_t$, relying on the source embedding $E_s$ representations:

$$g_{clp}(t_i, E_s, V_s, E_h, V_h) = \sum_{t_j \in V_t \cap V_s} E_s^{t_j} \cdot \alpha(E_h^{t_i}, E_h^{t_j})$$

\noindent where $\alpha(\cdot, \cdot)$ indicates a similarity score between two tokens in $E_h$. Here, we rely on the similarity function used by \citet{ostendorff2023efficient} computed as normalized cosine similarity.

\paragraph{SAVA} Mapping embedding representations between the embedding spaces of two different models using a linear model comes with theoretical justification. \citet{moschella2022relative} and \citet{maiorca2024latent} have shown that the embeddings of different models are related by a \textit{conformal translation}, or more generally, by a \textit{linear mapping} between such spaces. Inspired by the findings of \citet{maiorca2024latent} and by the intriguing effort of \citet{koto-etal-2021-indobertweet}, we propose a technique to perform vocabulary adaptation for generative models called \textit{Semantic Alignment Vocabulary Adaptation (SAVA)}. In our approach, we rely on a helper model embedding $E_h$ from an LLM and learn a linear mapping $\phi$ between $E_h \subseteq \mathbb{R}^m$ and $E_s \subseteq \mathbb{R}^n$. We train a single-layer Feed Forward Network (FFN) to map the helper embedding space onto the source embedding one: 

$$\phi \; : \; x \mapsto y \;\; | \; x \in \mathbb{R}^m , \; y \in \mathbb{R}^n, $$
\begin{equation}\label{eq:sava}
g_{sava}(t_i, E_h) = \phi(E_h^{t_i})
\end{equation}

The goal in training $\phi$ is to obtain a mapping between the representations of the tokens of the helper model and those of the source one. To train it, we use the tokens in the intersection $V_s \cap V_t$ since they have a representation according to both the source and the helper model, and we can train a linear map between the representations in $E_s$ and those in $E_h$. Then, as outlined from equation \ref{eq:sava} we use $\phi$ to map the tokens not present in the source vocabulary ($V_t \setminus V_s$) into the source embedding space. Therefore, our objective is to find:
\[
\phi(x) = W x + b,
\]
such that,
\[
\min_{W,\, b} \sum_{t_i \in V_s \cap V_t} \left\| W E_h^{t_i} + b - E_s^{t_i} \right\|^2.
\]
where $W \in \mathbb{R}^{n \times m}$ and $b \in \mathbb{R}^n$ are the parameters of our linear mapping.
More technical details about the training of the linear mapping are provided in Appendix \ref{app:sava_details}.

\subsection{Continual Training}

While re-initializing embeddings through vocabulary adaptation techniques enables zero-shot language modeling, the resulting language model often lacks proficiency in the new language. We address this by performing continual training on a mixture of source and target languages, which allows the model to retain performance in the source language while improving in the target language.

To achieve a robust comparison, we adapt pretrained LLMs to the target language using all the vocabulary adaptation heuristics discussed above. We also present results from continual training of the base model on the target language (LAPT). While less disruptive, this approach does not alter the vocabulary or tokenizer, preserving its fertility.

\section{Experimental Setup}\label{sec:exp_setup}

This section describes the setup of our experiments where we adapt two popular LLMs, specifically \mistral{} \citep{jiang2023mistral} and \llama{} \citep{dubey2024llama}. In the following subsections we report the settings used to do vocabulary adaptation, continual training and evaluation.

\subsection{Vocabulary Adaptation}

To adapt English models to the Italian language we rely on the \minervallms{} model family and its tokenizer \citep{orlando2024minerva}. The models of the \minervallms{} family are trained from scratch on an Italian-English dataset, i.e. CulturaX \citep{nguyen-etal-2024-culturax}. At the time of writing, three different models have been released, \minervas{}, \minervam{}, and \minerval{}, with the same tokenizer.

The \minervallms{} tokenizer shares 16,438 tokens with \mistral{} and 20,358 tokens with \llama{}. %
 For both \textit{CLP} and \textit{SAVA}, we use \minerval{} as the helper model.\footnote{We conduct ablation studies for the SAVA method, changing the number of tokens used to train $\phi$ and the size of the helper model. Some considerations are reported in Appendix \ref{app:sava_ablation}.}
Notably, as shown in \Cref{tab:param_count_w_adapt},  adapting a large model like \llama{} with \minervallms{} tokenizer significantly reduces the vocabulary size (by 75\%) and thus results in fewer parameters. The adapted \llama{} has 7.25B parameters compared to the original 8B, resulting in a 10\% reduction in model size.

As a further improvement, substituting the \mistral{} and \llama{} tokenizers with \minervallms{} one has a significant impact on the fertility in the Italian language. As shown in \Cref{tab:fertility}, the \minervallms{} tokenizer has on average 25\% of fertility gain compared to the \mistral{} tokenizer on two Italian text sources, CulturaX (CX) and Wikipedia (Wp). In the same setting, \llama{} improves its fertility up to 16\% on Italian text relying on the \minervallms{} tokenizer.

\begin{table}[t]
  \centering
  \adjustbox{max width=\linewidth}{%
  \begin{tabular}{lcc}
    \hline
    \textbf{Model} & \textbf{Num. Tokens} & \textbf{Num. Parameters} \\
    \hline
    Mistral-7B-v0.1 & 32000 & 7.24B \\
    Mistral-7B-v0.1 a.w. Minerva & 32768 & 7.25B \\
    \hline
    LLaMa-3-8B & 128256 & 8.03B \\
    LLaMa-3-8B a.w. Minerva & 32768 & 7.25B \\
    \hline
  \end{tabular}
  }
  \caption{Comparisons of model parameter counts and vocabulary size with and without adaptation (a.w. stands for \emph{adapted with}).}
  \label{tab:param_count_w_adapt}
\end{table}

\subsection{Continual Training}

To perform continual training we use CulturaX, a large-scale multilingual dataset that has been successfully used in large-scale continual training experiments on languages spoken within the European Union, including Italian.\footnote{\url{https://huggingface.co/occiglot/occiglot-7b-it-en-instruct}}
We aim to compare all methods on a fixed amount of compute budget, i.e. number of tokens. %
Due to a constrained computational budget, we decide to stop training after a threshold of 12B training tokens.

We subsample training data from the Italian and English splits of CulturaX to create a dataset composed of 75\% Italian tokens and 25\% English tokens, as proposed by \citet{csaki-etal-2024-sambalingo}. 

We use packing to fit all the tokens into sequences of a fixed length. The learning rate is fixed for all runs at $10^{-5}$.

For \mistral{}, training is done on 16 nodes on the Leonardo Supercomputer (each node uses 4 x 64 GB A100) maintaining a global batch size of 3072, and a sequence length of 2048.
For \llama{} we change the sequence length of the training data to 8192, and set the global batch size to 512. When training both models we do not freeze any parameter and let them all update. We perform continual training, allowing the models to process approximately 12 billion tokens. Specifically, we train \mistral{} for 2000 batches and \llama{} for 3000 batches. We use \emph{llm-foundry} for training\footnote{\url{https://github.com/mosaicml/llm-foundry}} and for the remaining hyper-parameters we use the default settings provided by the library. See \Cref{app:training_recs} for an estimation of the CO$_2$ cost of the experiments carried out in this work.

\begin{table}[t]
  \centering
  \adjustbox{max width=\linewidth}{%
  \begin{tabular}{lcccc}
    & \multicolumn{4}{c}{{\Large Fertility} $\downarrow$} \\
    \hline
    \textbf{Model} & \textbf{CX IT} & \textbf{CX EN} & \textbf{Wp IT} & \textbf{Wp EN} \\
    \hline
    Mistral-7B-v0.1 & 1.88 & 1.32 & 2.05 & 1.57 \\
    Minerva & \textbf{1.39} & 1.32 & \textbf{1.66} & 1.59 \\
    LLaMa-3-8B & 1.67 & \textbf{1.15} & 1.80 & \textbf{1.31} \\
    \hline
  \end{tabular}
  }
  \caption{Fertility of different tokenizers on CulturaX (CX) and Wikipedia (Wp).}
  \label{tab:fertility}
\end{table}

\subsection{Evaluation}

\begin{table*}[t]
  \centering
  \adjustbox{max width=0.9\linewidth}{%
  \begin{tabular}{lcccccc|c}
    \hline
    \textbf{Model} & \textbf{Hellaswag} & \textbf{MMLU} & \textbf{Arc Easy} & \textbf{PIQA} & \textbf{SCIQ} &   \textbf{BOOLQ} & \textbf{AVG}\\
    \hline
    \rowcolor{gray!20}Mistral-7B-v0.1 & $56.50_{\pm0.49}$ & $47.42_{\pm0.42}$ & $61.67_{\pm1.01}$ & $67.24_{\pm1.14}$ & $84.75_{\pm1.16}$ & $75.01_{\pm0.75}$ & 65.43 \\\hline
    & \multicolumn{7}{c}{200 Training Steps} \\\hline
    Random & $55.60_{\pm0.49}$ & $42.48_{\pm0.42}$ & $57.92_{\pm1.02}$ & $68.05_{\pm1.16}$ & $75.46_{\pm1.39}$ & $72.29_{\pm0.78}$ & 61.96 \\
    FVT & $\underline{56.34}_{\pm0.49}$ & $\textbf{44.28}_{\pm0.42}$ & $\underline{60.42}_{\pm1.01}$ & $\textbf{69.90}_{\pm1.14}$ & $\textbf{80.48}_{\pm1.28}$ & $\textbf{74.52}_{\pm0.76}$ & \textbf{64.32} \\
    CLP & $54.74_{\pm0.49}$ & $42.50_{\pm0.42}$ & $57.62_{\pm1.02}$ & $67.74_{\pm1.16}$ & $76.82_{\pm1.36}$ & $68.07_{\pm0.81}$ & 61.24 \\
    SAVA & $\textbf{56.73}_{\pm0.49}$ & $\underline{44.23}_{\pm0.42}$ & $\textbf{60.90}_{\pm1.01}$ & $\underline{69.72}_{\pm1.14}$ & $\underline{79.22}_{\pm1.31}$ & $\underline{73.30}_{\pm0.77}$ & \underline{64.01} \\\hline
    LAPT & $58.29_{\pm0.49}$ & $49.31_{\pm0.42}$ & $63.00_{\pm1.00}$ & $69.84_{\pm1.14}$ & $84.13_{\pm1.18}$ & $75.07_{\pm0.75}$ & 66.60 \\\hline
    \rowcolor{cyan!20}& \multicolumn{7}{c}{2000 Training Steps} \\\hline
    \rowcolor{cyan!20}Random & $58.43_{\pm0.49}$ & $46.95_{\pm0.42}$ & $62.87_{\pm1.00}$ & $\underline{71.39}_{\pm1.12}$ & $81.62_{\pm1.25}$ & $72.47_{\pm0.78}$ & 65.62 \\
    \rowcolor{cyan!20}FVT & $59.00_{\pm0.49}$ & $\underline{47.35}_{\pm0.42}$ & $\textbf{63.52}_{\pm0.99}$ & $\textbf{71.51}_{\pm1.12}$ & $\textbf{84.55}_{\pm1.16}$ & $75.74_{\pm0.74}$ & 66.94 \\
    \rowcolor{cyan!20}CLP & $\underline{59.21}_{\pm0.49}$ & $47.10_{\pm0.42}$ & $\underline{63.47}_{\pm0.99}$ & $70.77_{\pm1.13}$ & $84.44_{\pm1.17}$ & $\textbf{76.75}_{\pm0.73}$ & \underline{66.95} \\
    \rowcolor{cyan!20}SAVA & $\textbf{59.41}_{\pm0.49}$ & $\textbf{47.57}_{\pm0.42}$ & $63.39_{\pm0.99}$ & $71.02_{\pm1.12}$ & $\textbf{84.55}_{\pm1.16}$ & $\underline{76.02}_{\pm0.74}$ & \textbf{66.99} \\\hline
    \rowcolor{cyan!20}LAPT & $60.51_{\pm0.48}$ & $46.63_{\pm0.42}$ & $64.99_{\pm0.99}$ & $71.21_{\pm1.12}$ & $85.90_{\pm1.12}$ & $76.17_{\pm0.74}$ & 67.56 \\\hline
  \end{tabular}
  }
  \caption{0-shot results over \textbf{Italian} translated benchmarks for \mistral{} adapted models.}
  \label{tab:zero_shot_it_mistral}
\end{table*}

To evaluate our models we rely on the LM-Evaluation-Harness library \citep{eval-harness}, for multiple-choice (MC) benchmarks, using the perplexity evaluation method. 
As MC benchmarks, we use the translated section of ITA-Bench \citep{moroni2024ita}, a suite of benchmarks automatically translated from English to Italian.

During continual training we evaluate our models every 200 batches for \mistral{} and 300 batches for \llama{} in a 0-shot scenario; in this way, each subsequent checkpoint is evaluated consistently on the same number of tokens. 
To assess the reasoning capabilities of the adapted models, we use a variety of benchmarks: \textbf{MMLU} \citep{hendryckstest2021}, \textbf{BOOLQ} \citep{clark-etal-2019-boolq}, \textbf{ARC-easy} \citep{allenai:arc}, \textbf{PIQA} \citep{Bisk2020}, \textbf{SciQ} \citep{welbl-etal-2017-crowdsourcing}, and \textbf{Hellaswag} \citep{zellers-etal-2019-Hellaswag}.

We also measure the model performance on generative tasks, focusing on two tasks: automatic translation, \textbf{FLoRes} benchmark \citep{nllb2022}, and question answering, \textbf{SQuAD-it} \citep{10.1007/978-3-030-03840-3_29}, a version of \textbf{SQuAD} \citep{rajpurkar-etal-2016-squad} automatically translated into Italian. We used \texttt{vLLM} \citep{kwon2023efficient} as our generation pipeline. More details related to the generation techniques can be found in Appendix \ref{app:generation}.

\begin{figure}[t]
  \includegraphics[width=\columnwidth]{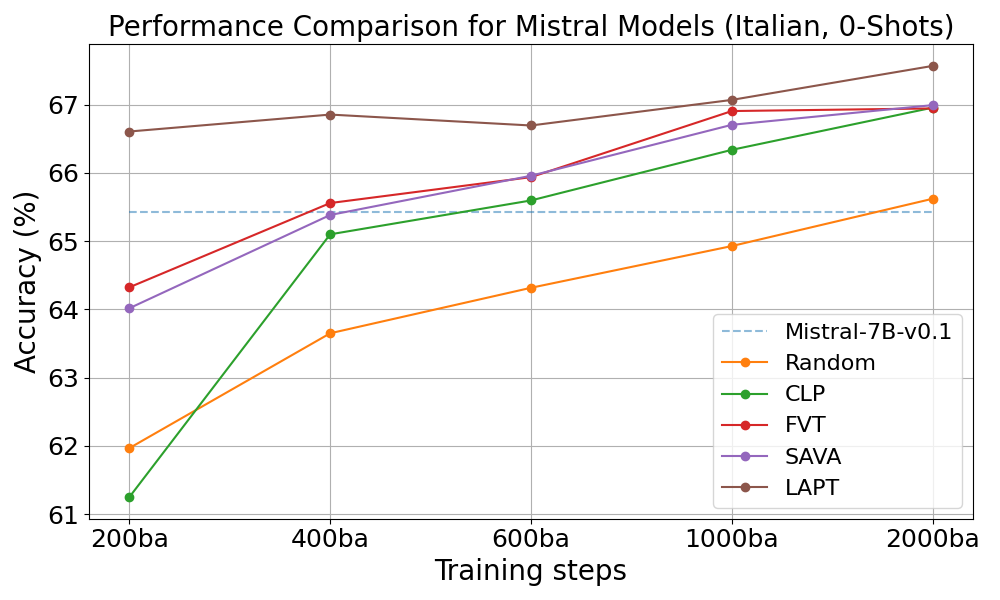}
  \caption{Average performance of \mistral{} based models during training on \textbf{Italian} translated benchmarks. The average was calculated over six datasets.}
  \label{fig:mistral_italian_average}
\end{figure}

\section{Results}

In this section, we discuss the results obtained from evaluating the adapted models. We begin by examining the scores on multiple-choice benchmarks, followed by a separate analysis of performance on generative benchmarks, specifically \textit{FLoRes} and \textit{SQuAD-it}. In this and subsequent sections we indicate the continual training of the base model without vocabulary adaptation by the \textit{LAPT} acronym.

\begin{table*}[t]
  \centering
  \adjustbox{max width=0.9\linewidth}{%
  \begin{tabular}{lcccccc|c}
    \hline
    \textbf{Model} & \textbf{Hellaswag} & \textbf{MMLU} & \textbf{Arc Easy} & \textbf{PIQA} & \textbf{SCIQ} & \textbf{BOOLQ} & \textbf{AVG} \\
    \hline
    \rowcolor{gray!20}LLaMa-3.1-8B & $57.97_{\pm0.49}$ & $54.28_{\pm0.42}$ & $60.46_{\pm1.01}$ & $68.54_{\pm1.15}$ & $82.77_{\pm1.22}$ & $74.52_{\pm0.76}$ & 66.42 \\\hline
    & \multicolumn{7}{c}{300 Training Steps} \\\hline
    FVT & $\textbf{55.61}_{\pm0.49}$ & $\textbf{50.24}_{\pm0.42}$ & $59.38_{\pm1.01}$ & $\textbf{66.99}_{\pm1.17}$ & $80.68_{\pm1.27}$ & $70.00_{\pm0.80}$ & 63.81 \\
    SAVA & $55.48_{\pm0.49}$ & $49.26_{\pm0.42}$ & $\textbf{59.77}_{\pm1.01}$ & $66.62_{\pm1.17}$ & $\textbf{81.31}_{\pm1.26}$ & $\textbf{74.43}_{\pm0.76}$ & $\textbf{64.48}$ \\\hline
    LAPT & $57.92_{\pm0.49}$ & $53.10_{\pm0.42}$ & $61.32_{\pm1.01}$ & $68.97_{\pm1.15}$ & $82.56_{\pm1.22}$ & $72.20_{\pm0.78}$ & 66.01 \\\hline
    \rowcolor{cyan!20}& \multicolumn{7}{c}{3000 Training Steps} \\\hline
    \rowcolor{cyan!20}FVT & $\textbf{58.44}_{\pm0.49}$ & $\textbf{51.47}_{\pm0.42}$ & $62.70_{\pm1.00}$ & $69.53_{\pm1.14}$ & $\textbf{83.29}_{\pm1.20}$ & $69.35_{\pm0.80}$ & 65.79 \\
    \rowcolor{cyan!20}SAVA & $57.82_{\pm0.49}$ & $51.08_{\pm0.42}$ & $\textbf{63.17}_{\pm1.00}$ & $\textbf{69.78}_{\pm1.14}$ & $81.73_{\pm1.24}$ & $\textbf{74.15}_{\pm0.76}$ & $\textbf{66.29}$ \\
    \hline
    \rowcolor{cyan!20}LAPT & $59.35_{\pm0.49}$ & $52.94_{\pm0.42}$ & $62.96_{\pm1.00}$ & $69.72_{\pm1.14}$ & $82.98_{\pm1.21}$ & $71.77_{\pm0.78}$ & 66.62 \\\hline
  \end{tabular}
  }
  \caption{0-shot results over \textbf{Italian} translated benchmarks for \llama{} adapted models.}
  \label{tab:zero_shot_it_llama}
\end{table*}

\subsection{Multi-choice Setting}

\subsubsection{Italian Results}

We report results on Italian benchmarks for \mistral{} after 200 and 2000 batches in \Cref{tab:zero_shot_it_mistral}. From the table we can see that the adapted models reach over random-chance performance at the beginning of training (200-step setting), with \textit{FVT} and \textit{SAVA} achieving higher performance compared to other methods (\textit{CLP} and \textit{Random}). All the vocabulary adaptation heuristics perform worse compared to the \textit{LAPT} technique, which is expected since \textit{LAPT} does not apply any disruptive architectural change to the model. Looking at the results at 2000 batches, we can see that all the adapted models surpass the scores of the base model, and the performance gap with \textit{LAPT} becomes low. Even in this setting, \textit{SAVA} and \textit{FVT} perform well, while \textit{Random} lags behind.

In \Cref{fig:mistral_italian_average}, we present the average scores across the six Italian tasks. \textit{SAVA} and \textit{FVT} consistently achieve higher overall scores throughout the training process, with a more pronounced advantage in the early stages. This highlights the influence of the chosen heuristic, particularly immediately after the vocabulary substitution. \textit{SAVA} and \textit{FVT} achieve results at 400 batches that are comparable to those of the \textit{Random} approach at the end of training, thereby reducing total training time by approximately 80\%.

\begin{figure}[t]
  \includegraphics[width=\columnwidth]{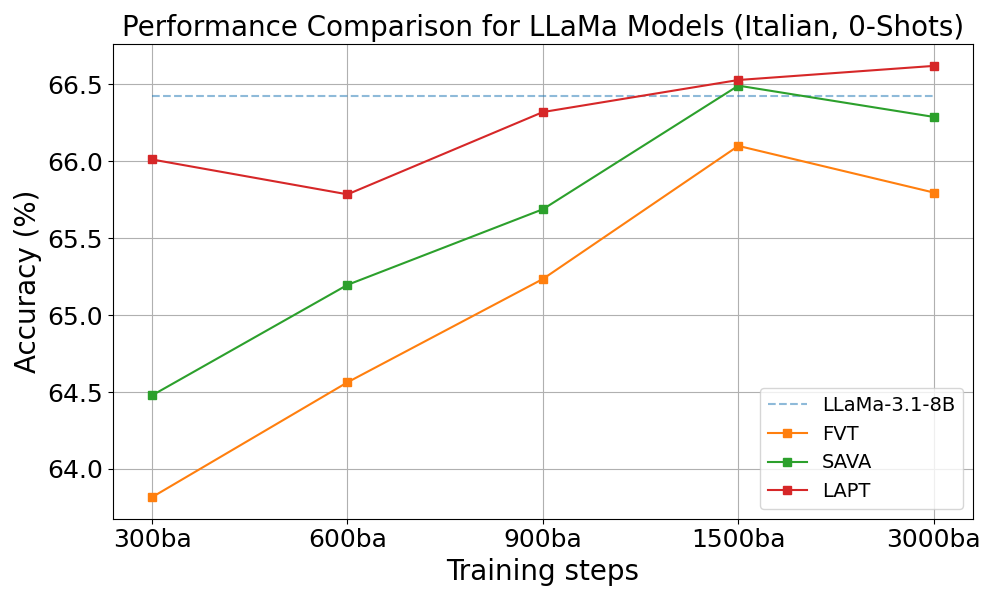}
  \caption{Average performance of \llama{} based models during training on \textbf{Italian} translated benchmarks. The average was calculated over six datasets.}
  \label{fig:llama_italian_average}
\end{figure}

In the case of \llama{}, \Cref{tab:zero_shot_it_llama} reports the scores of the adapted models, after 300 and 3000 batches. We show that \textit{FVT} and \textit{SAVA} maintain comparable performance, except for \textit{BOOLQ} where \textit{SAVA} showcases better scores, $+4\%$, even in comparison to the \textit{LAPT} setting.
Compared to the adapted models, the \llama{} model remains a strong baseline on Italian tasks. Still in that setting, we further narrow the performance gap with the \textit{LAPT} model using both vocabulary adaptation heuristics. In \Cref{fig:llama_italian_average}, we report the average scores on Italian tasks and observe a constant improvement through the training steps.

\subsubsection{English Results}

Including English in the evaluation allows us to assess whether performance on the source language is preserved during continual training, for both \mistral{} and \llama{}. As mentioned in \Cref{sec:exp_setup}, we train on mainly Italian data and a smaller portion of English (25\% of the total).

\begin{figure}[t]
  \includegraphics[width=\columnwidth]{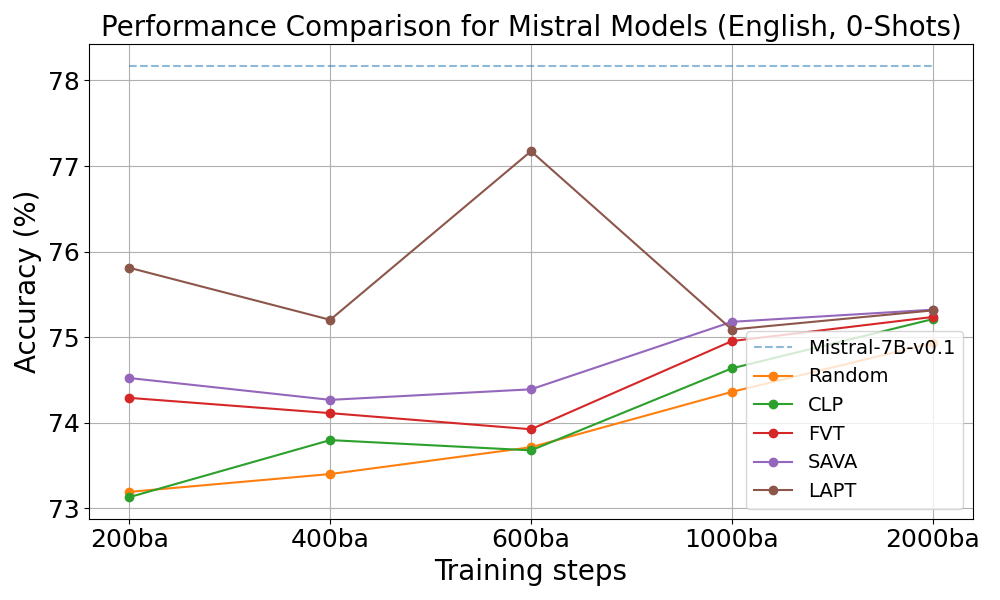}
  \caption{Average performance of \mistral{} based models during training on \textbf{English} benchmarks. The average was calculated over six datasets.}
  \label{fig:mistral_english_average}
\end{figure}

\Cref{fig:mistral_english_average} reports the average scores during training on English texts for \mistral{}. We can see that all trained models reach a comparable average score at the end of the adaptation process. All the adapted models diminish in performance compared to the base one in the English language.

\begin{figure}[t]
  \includegraphics[width=\columnwidth]{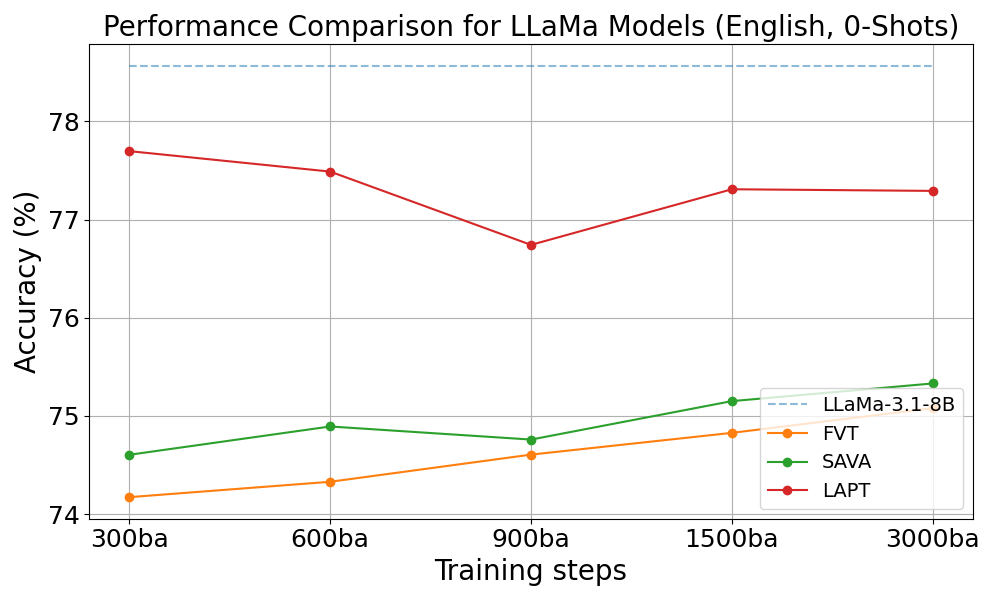}
  \caption{Average performance of \llama{} based models during training on \textbf{English} benchmarks. The average was calculated over six datasets.}
  \label{fig:llama_english_average}
\end{figure}

\Cref{fig:llama_english_average} reports the average scores of \llama{} models on English benchmarks during training. In this setting, \textit{LAPT} maintains higher performance on average; intuitively, this could be attributed to the larger vocabulary of \llama{} (75\% bigger), which enables better performance during language adaptation, avoiding catastrophic forgetting of the source language.

For both models the \textit{SAVA} approach leads the model to achieve slightly higher performance in the source language.
Appendix \ref{app:ensligh_results} reports more detailed results of evaluation over English benchmarks.

\subsection{Generative Setting}
Multi-choice benchmarking based on perplexity scoring has its own limitations \citep{wang-etal-2024-answer-c}. To further test our models, we evaluate them on two generative tasks: Machine Translation (MT), IT-EN and EN-IT, and Italian Question Answering.

We report COMET-22 \citep{rei-etal-2022-comet} for the MT benchmark and RougeL \citep{lin-2004-rouge} for the Question Answering task.

Looking at the MT results, in \Cref{tab:gen_mistral_res}, we observe that the adapted \mistral{} models achieve excellent performance, outperforming those of the base model. The vocabulary adapted models reach very good results in the English-to-Italian direction, where generation of Italian text is involved. Our findings indicate that \textit{SAVA} and \textit{FVT} emerge as the most effective vocabulary adaptation heuristics in this context. As shown in \Cref{tab:gen_llama_res}, a similar trend is observed with \llama{}, where adapted models perform competitively with the base model, while \textit{SAVA} and \textit{FVT} reach the same performance as those of \textit{LAPT}.

Regarding the results in the SQuAD-it task, \Cref{tab:gen_mistral_res,tab:gen_llama_res} show that \textit{SAVA} attains very good performance, beating other heuristics and the \textit{LAPT} approach for both model types, reaching inline performance equal to that of the base model for \llama{}.

\subsection{Training Loss}

\begin{table}[t]
  \centering
  \adjustbox{max width=\linewidth}{%
  \begin{tabular}{lcc|c}
     & \multicolumn{2}{c}{FLoRes} & \multicolumn{1}{c}{SQuAD-it} \\
     \cline{2-4}
     \hline
    \textbf{Model} & \textbf{EN-IT} & \textbf{IT-EN} & \textbf{RL} \\
    \hline
    \rowcolor{gray!20}\mistral{} & 86.57 & 87.75 & 68.92 \\
    \hline
     & \multicolumn{3}{c}{200 Training Steps} \\
    \hline
    Random  & 86.67 & 87.37 & 62.1 \\
    FVT & \underline{87.08} & \underline{87.55} & \underline{65.47} \\
    CLP & 86.58 & 87.31 & 64.25 \\
    SAVA & \textbf{87.30} & \textbf{87.59} & \textbf{65.66} \\
    \hline
    LAPT & 87.41 & 87.92 & 67.35 \\
    \hline
    \rowcolor{cyan!20}& \multicolumn{3}{c}{2000 Training Steps} \\\hline
    \rowcolor{cyan!20}Random  & 88.01 & \textbf{87.92} & 64.83 \\
    \rowcolor{cyan!20}FVT & \underline{88.29} & \underline{87.90} & \underline{66.18} \\
    \rowcolor{cyan!20}CLP & 88.21 & 87.79 & 65.99 \\
    \rowcolor{cyan!20}SAVA & \textbf{88.31} & 87.87 & \textbf{67.20} \\
    \hline
    \rowcolor{cyan!20}LAPT & 88.13 & 88.02 & 66.92 \\
    \hline
  \end{tabular}
  }
  \caption{5-shot results for \mistral{} of FLoRes where COMET-22 is reported and 2-shot results for SQuAD-it where RougeL is reported.}
  \label{tab:gen_mistral_res}
\end{table}

\begin{table}[t]
  \centering
  \adjustbox{max width=\linewidth}{%
  \begin{tabular}{lcc|c}
     & \multicolumn{2}{c}{FLoRes} & \multicolumn{1}{c}{SQuAD-it} \\
     \cline{2-4}
     \hline
    \textbf{Model} & \textbf{EN-IT} & \textbf{IT-EN} & \textbf{RL} \\
    \hline
    \rowcolor{gray!20}\llama{} & 87.59 & 88.08 & 69.21\\
    \hline
     & \multicolumn{3}{c}{300 Training Steps} \\
    \hline
    FVT  & 87.32 & \textbf{87.65} & 68.54 \\
    SAVA & \textbf{87.39} & 87.58 & \textbf{68.70} \\
    \hline
    LAPT & 87.82 & 87.95 & 67.91 \\
    \hline
    \rowcolor{cyan!20}& \multicolumn{3}{c}{3000 Training Steps} \\\hline
    \rowcolor{cyan!20}FVT  & 88.05 & 88.02 & 68.84 \\
    \rowcolor{cyan!20}SAVA & \textbf{88.12} & \textbf{88.04} & \textbf{69.05} \\
    \hline
    \rowcolor{cyan!20}LAPT & 88.11 & 88.05 & 66.69 \\
    \hline
  \end{tabular}
  }
  \caption{5-shot results for \llama{} of FLoRes where COMET-22 is reported and 2-shot results for SQuAD-it where RougeL is reported.}
  \label{tab:gen_llama_res}
\end{table}

\begin{figure}[t]
  \includegraphics[width=\columnwidth]{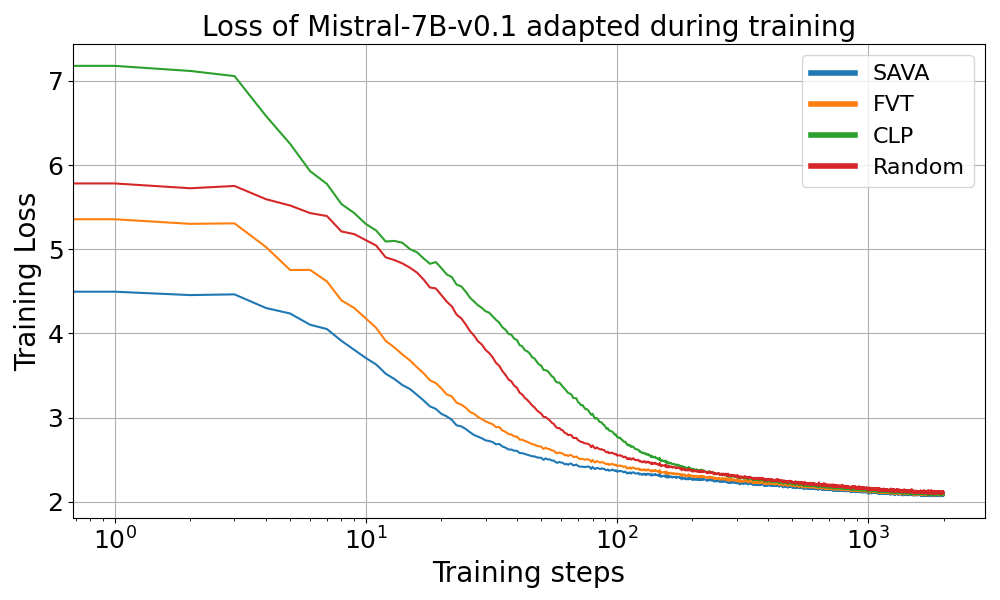}
  \caption{Loss during continual training of \mistral{} models.}
  \label{fig:training_loss_mistral}
\end{figure}

\begin{figure}[t]
  \includegraphics[width=\columnwidth]{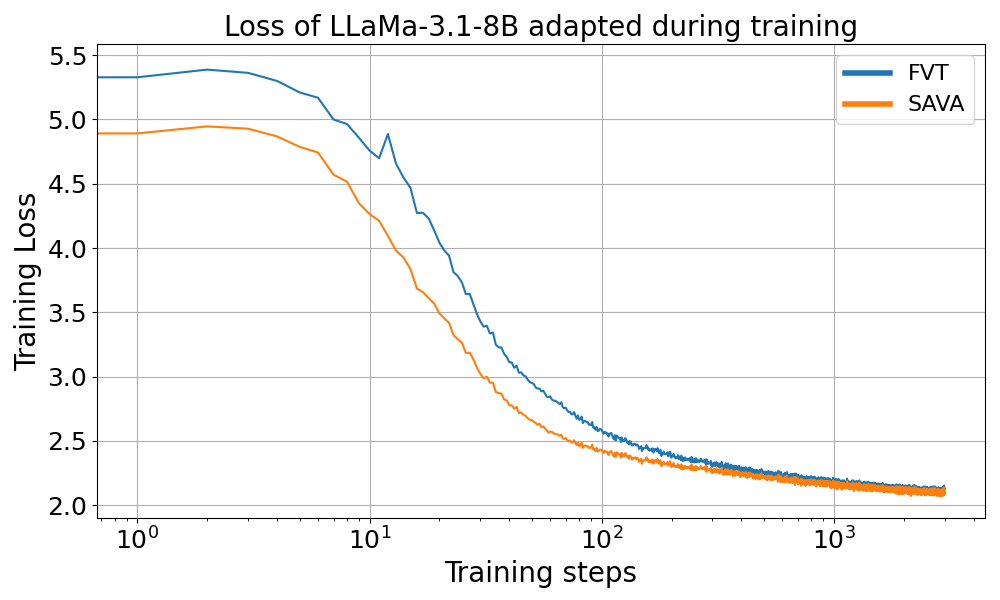}
  \caption{Loss during continual training of \llama{} models.}
  \label{fig:training_loss_llama}
\end{figure}

Important observations can be made concerning the loss trajectories. \Cref{fig:training_loss_mistral} reports the \mistral{} plots, and we can notice significant differences between the various heuristics in the early stages of the training. The \textit{SAVA}-model emerges as the better-adapted one, right from the start, particularly when compared to the \textit{CLP} and \textit{Random} models. Notably, \textit{CLP} appears to lag behind \textit{Random} initially. Looking at \llama{} losses, in \Cref{fig:training_loss_llama} we can see that the two heuristics exhibit similar trajectories, although \textit{SAVA} still achieves a lower loss from the outset.

\section{Differences in the Embedding Structure}
To better understand the impact of different vocabulary adaptation techniques, we analyze similarities in intra-model and inter-model embedding spaces. Specifically, we examine how different adaptations influence the structural alignment of embeddings in comparison to a reference model (intra-model similarity) and how the embedding spaces of different adapted models compare to each other (inter-model similarity).

To measure the similarity between two embedding spaces, we rely on the technique introduced by \citet{moschella2022relative}. Specifically, we randomly select 128 non-prefix tokens and 128 prefix tokens from \(V_t\) to compute relative embedding representations, resulting in a total of 256 anchor tokens.\footnote{Non-prefix tokens refer to complete words or sub-words that do not start a sequence (e.g., “cat” in “concatenate”), while prefix tokens initiate a word or sub-word.} For each model, we then adjust the representation of each token relative to these anchors, calculating each dimension as the projection onto the selected anchors. Subsequently, we compute the cosine similarity based on this relative representation across the models and average the results to obtain an overall similarity score between the two distinct models.

\paragraph{Intra-model similarity}
Intuitively, a well-adapted model should align with \minerval{}, as it serves as a strong reference for the target language. Similarly to our setting, \minerval{} is pretrained on balanced Italian-English data from CulturaX.
In \Cref{tab:sim_mistral_llama}, we present the similarity scores between adapted models and \minerval{}. Notably, \textit{CLP} and \textit{SAVA} achieve higher similarity scores than other approaches. This outcome is to be expected, as both \textit{CLP} and \textit{SAVA} leverage \minerval{}'s embedding space. Interestingly, \textit{SAVA} not only attains a structure that is more similar to that of \minerval{} ($+3.7$), but also demonstrates superior performance, as was also the case in previous sections.

\begin{table}[t]
  \centering
  \adjustbox{max width=\linewidth}{%
  \begin{tabular}{lcc|cc}
     & \multicolumn{2}{c}{\mistral{}} & \multicolumn{2}{c}{\llama{}} \\
    \textbf{Model} & \textbf{@0ba} & \textbf{@2000ba} & \textbf{@0ba} & \textbf{@3000ba}\\
    \hline
    Random & 29.68 & 31.67 & - & - \\
    FVT & 33.65 & 35.30 & 33.23 & 33.49 \\
    CLP & \underline{41.10} & \underline{42.84} & - & - \\
    SAVA & \textbf{44.81} & \textbf{45.33} & \textbf{41.84} & \textbf{42.02}  \\
    \hline
  \end{tabular}
  }
  \caption{Similarity scores between \mistral{} adapted models and \minerval{} (left) and between \llama{} and \minerval{} (right) at the beginning and at the end of the training.}
  \label{tab:sim_mistral_llama}
\end{table}

\begin{figure}[t]
\centering
  \includegraphics[width=0.85\columnwidth]{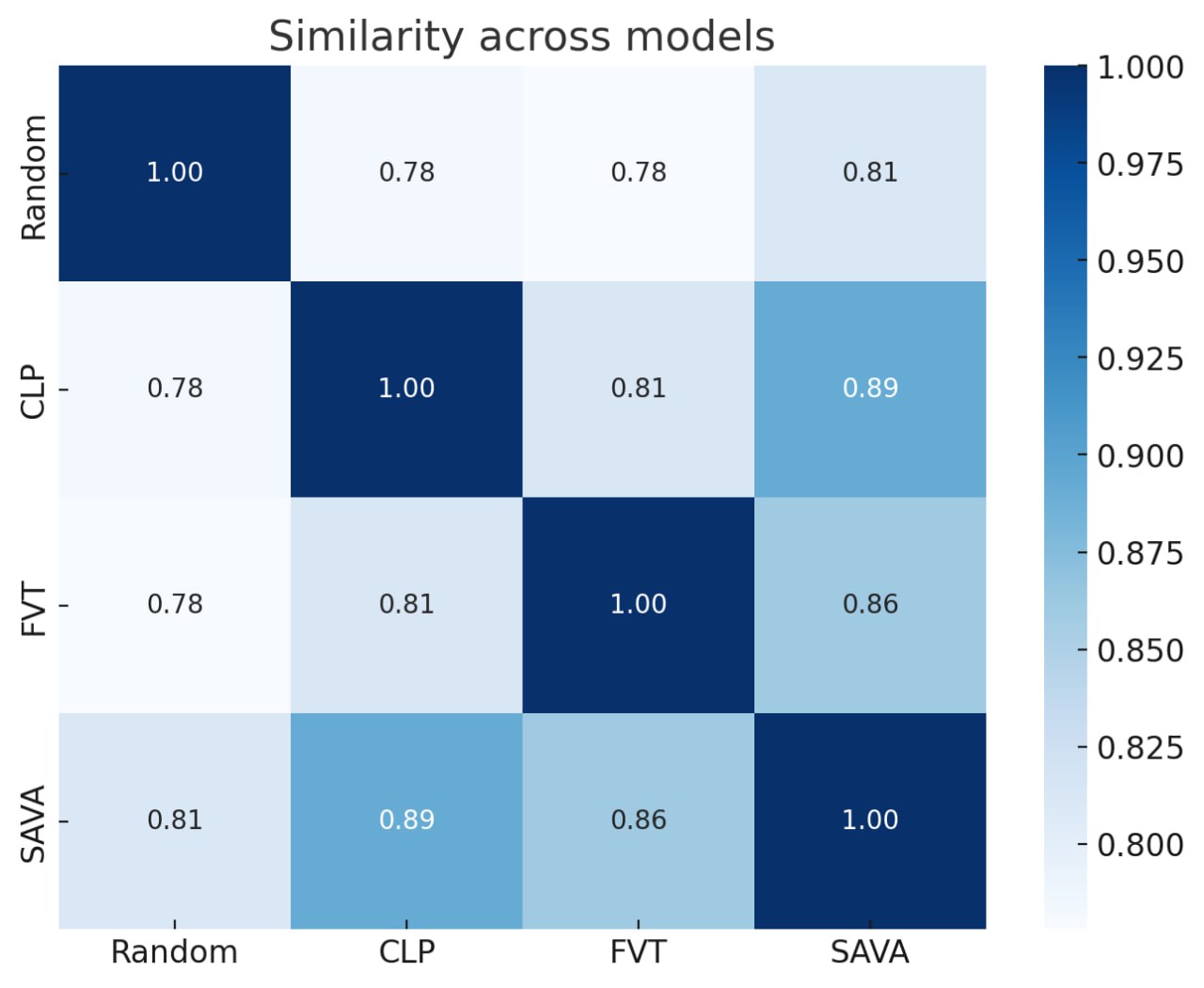}
  \caption{Similarity across models after continual training on 12B tokens.}
  \label{fig:sim_cross_models}
\end{figure}

\paragraph{Inter-model similarity}
To gain deeper insights into the differences in learned embedding structures, \Cref{fig:sim_cross_models} presents the similarity scores between \mistral{} variants adapted using the specified techniques. We compare the models at the end of the continual training. The analysis shows a high similarity between models, but differences of up to 10\% in the relative representations reveal structural variations in the encoded information. This analysis suggests that,  even after intensive training, the adapted models do not converge to the same representation.

\section{Conclusions}
In this work, we extensively explored various techniques to adapt English-focused LLMs, i.e. \mistral{} and \llama{}, to the Italian language. We introduced a novel heuristic called \textit{SAVA} which leverages the embedding structure of a smaller, native Italian Language Model, \minerval{}. We discovered that adapting the vocabulary of English LLMs leads to significant improvements in language encoding, reducing the number of generated tokens by 25\% for \mistral{} and 16\% for \llama{}. Regarding \llama{} we pruned nearly 1 billion parameters by optimizing its vocabulary, removing approximately 75\% of the original tokens. Our evaluation revealed performance differences across the vocabulary adaptation heuristics, by means of a thorough analysis during the continual training phase. We show that linguistic capabilities can be restored with relatively few training steps—\mistral{} reached base model performance after processing 2 billion tokens. Additionally, the \textit{SAVA} heuristic demonstrated strong performance on downstream tasks, with \textit{SAVA}-adapted models reaching faster convergence during continual training. Furthermore, the embedding structure of \textit{SAVA} exhibited closer alignment with the helper model compared to other analyzed heuristics. 

This work opens several research directions. One key area of interest will be to evaluate how the \textit{SAVA} approach scales across languages, particularly in mid- and low-resource settings. Understanding how different heuristics perform with a small number of continual training steps in such scenarios is crucial. Additionally, since \textit{Minerva-7B} was not available at the time of writing, a logical next step would be to utilize it as a helper model.

\section{Limitations}
We investigated the adaptation of English-first LLMs to the Italian language with a focus on adapting the vocabulary and the tokenizer to match the performance of continually trained models while achieving lower fertility and thus higher efficiency in the target language. 

We limited our training data to the CulturaX dataset, which consists of cleaned web-crawled data. Incorporating higher-quality datasets could improve the models' performance in the target language.

We limited our analysis to two distinct decoder-only Large Language Models: \mistral{} and \llama{}. For a more comprehensive study, additional English-first models could be tested. However, the aforementioned two models are among the best performing ones in their parameter count. Furthermore, we chose to focus on just two models due to the extensive continual training we had to perform, as such training requires considerable computational resources.

We evaluated the adapted models on automatically translated datasets for multiple-choice tasks and open-ended question answering. Specifically, Hellaswag, MMLU, Arc Easy, PIQA, SCIQ, and BOOLQ were translated using \textit{Tower-Instruct-v0.2}, an open-source solution for automatic translation that, at the time of writing, represents the state of the art in open Machine Translation models. For generative tasks, SQuAD-it was translated using a semi-automatic approach.

We acknowledge that relying on automatically translated benchmarks may have introduced some noise, potentially obscuring certain abilities or issues in the models’ comprehension of Italian texts. This limitation was beyond our capabilities to resolve since no well-structured Italian native benchmarks exist. Another limitation was using only two generative benchmarks, where we observed slightly different results for the adapted models. In the generative setting, SAVA generally outperformed other methods, while LAPT models did not consistently deliver the best average performance on downstream tasks.

Future work should aim to explore the capabilities of vocabulary-adapted models in generative tasks and investigate how a model's fertility over target language influences downstream performance.

\section{Ethics Statement}
We primarily conduct experiments in the Italian language. This approach is aimed at addressing the practical challenges of working with Italian, a language that is underrepresented in the NLP field.
Our continual training is performed on data collected from open web sources, specifically through the CulturaX dataset. Since large-scale datasets used for pretraining can include personal and sensitive information, it is crucial to carefully assess such content before deploying models in real-world applications.
Another key consideration is the use of existing monolingual or multilingual models as starting points, rather than training new models from scratch. This can introduce biases from the original pretraining data, potentially causing the model to reflect behaviors and cultural influences from other languages rather than those of the target language community.

\section*{Acknowledgments}
Edoardo Barba and Alessio Miaschi are fully funded by the PNRR MUR project \href{https://fondazione-fair.it/}{PE0000013-FAIR}. Roberto Navigli and Felice Dell'Orletta acknowledge the support of the PNRR MUR project \href{https://fondazione-fair.it/}{PE0000013-FAIR}. Partially financed by the European Union - NextGenerationEU through the Italian Ministry of University and Research under PNRR - PRIN 2022 (2022EPTPJ9) "WEMB: Word Embeddings from Cognitive Linguistics to Language Engineering and back" and by the PNRR project ITSERR (CUP B53C22001770006). We acknowledge the support of the ISCRA project TRAVEL (HP10CY9V7K) for awarding access to the LEONARDO supercomputer, owned by the EuroHPC Joint Undertaking, hosted by CINECA (Italy) and thank Giuseppe Fiameni for his support.

\bibliography{anthology,custom}

\appendix

\section{SAVA Training of the mapping function}
\label{app:sava_details}

To implement the SAVA methods, we first need to train the linear mapping function, \(\phi\). For this, we use the \texttt{SGDAffineAligner} method provided in the \textbf{latentis} library\footnote{\url{https://github.com/Flegyas/latentis}}.

After collecting the token representation pairs from the intersection, we train the linear mapping using the \texttt{ADAM} optimizer with \texttt{MSE Loss}, setting the learning rate to \(10^{-3}\) and running the optimization for 1000 steps.

To enhance training stability, we first apply \textit{standard scaling} and \textit{L2 normalization} to the token representations before learning \(\phi\). After training, we apply the inverse scaling to restore the original distribution before incorporating the results into the adapted model.

\section{Ablation experiments on the SAVA method}
\label{app:sava_ablation}
\begin{figure}[t]
  \includegraphics[width=\columnwidth]{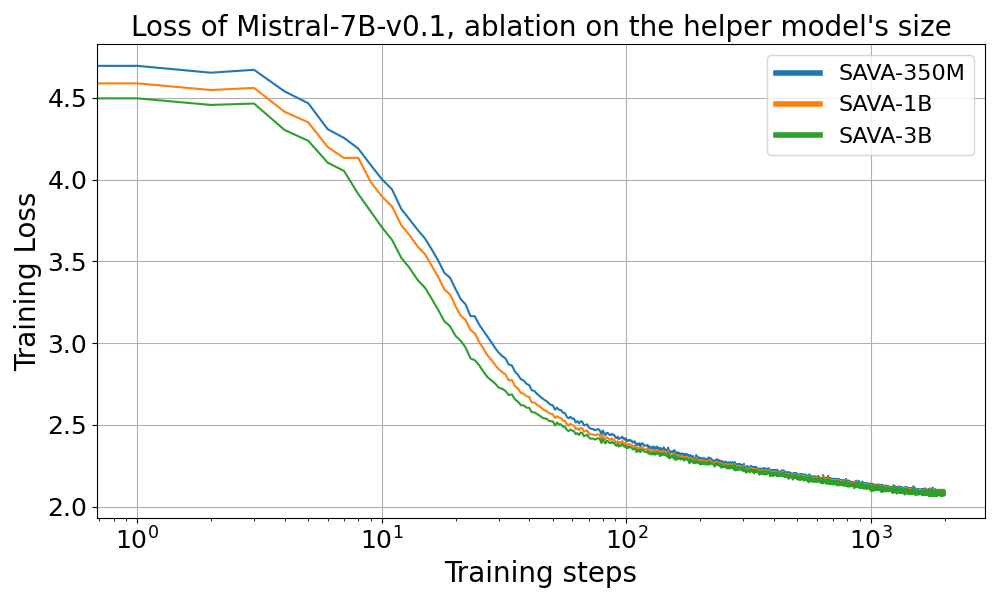}
  \caption{Loss during continual training of Mistral models.}
  \label{fig:training_loss_model_abl}
\end{figure}

\begin{figure}[t]
  \includegraphics[width=\columnwidth]{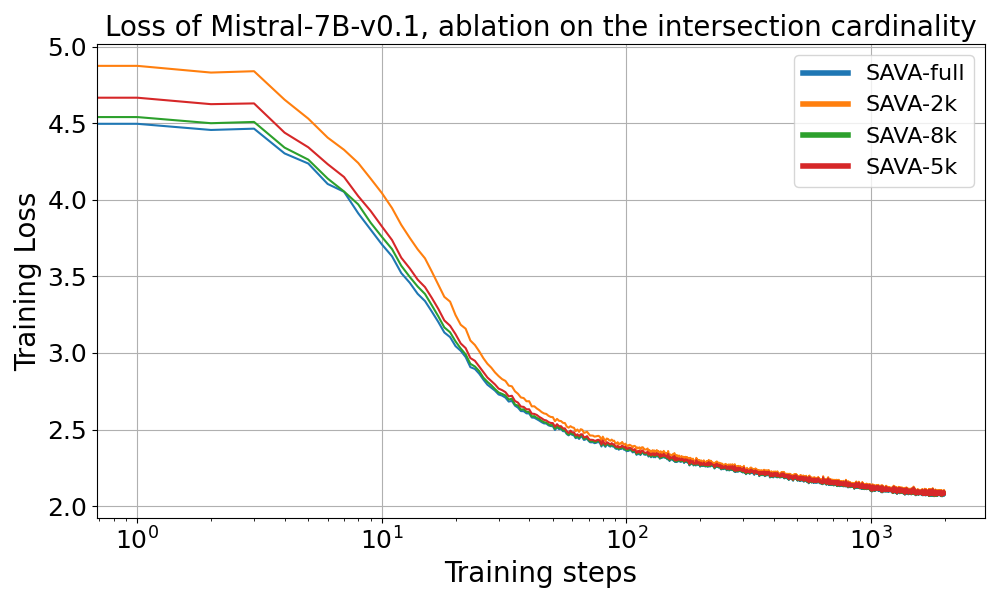}
  \caption{Loss during continual training of Mistral models.}
  \label{fig:training_loss_anchors_abl}
\end{figure}

In this section we analyze some ablation studies over the \textit{SAVA} method. We analyzed the impact on the helper model's size, using the two smaller models of Minerva's family, \minervas{} and \minervam{}, which have, respectively, 350M and 1B parameters. In \Cref{fig:training_loss_model_abl} the training loss of \mistral{} adapted using \textit{SAVA} with different helper models is reported. From the plot we can see that the dimension of the helper model does not have a huge impact on the loss trajectory.
An orthogonal experiment was conducted to ablate the number of tokens used to learn the mapping $\phi$, in \Cref{fig:training_loss_anchors_abl} the loss for \mistral{} adapted with \textit{SAVA} relying on different number of tokens in $V_t \cap V_s$, over \minerval{} is reported. We observe that using more tokens leads to a faster convergence of the training loss. From the plots we can see that reducing the number of tokens has a greater impact than reducing the model size, especially for the setting with two thousand tokens.

\section{Training Resources and Environmental Impact}
\label{app:training_recs}

Experiments were conducted using the LEONARDO Italian Supercomputer, which has a carbon efficiency of 0.432 kgCO$_2$eq/kWh. A cumulative of 50000 hours of computation was performed on hardware of type A100 SXM4 80 GB (TDP of 400W).

Total emissions are estimated to have been 8640 kgCO$_2$eq of which 0 percent were directly offset. These emissions were split roughly into 95\% for continual training and 5\% for evaluation.

This is an approximate estimate since the computation was done on LEONARDO custom hardware which is not available in the tool used for the estimation \href{https://mlco2.github.io/impact#compute}.

\section{Generation setting}
\label{app:generation}
We tested our adapted models on two downstream tasks in a generative setting, machine translation and question answering. We tested the models in few-shot setting relying on the in-context capabilities of evaluated models, without any fine-tuning step to the specific task. We relied on the vLLM library \cite{kwon2023efficient} to afford prompting generation, specifically we changed the default parameters with \texttt{temperature=0} and \texttt{max\_tokens=512}.

After a comprehensive number of trials we noticed that the prompting strategy had a huge impact, while the order between the models remained unchanged. We report the prompts used for FLoRes and SQuAD-it tasks in \Cref{tab:FLoRes_prompt,tab:SQuAD_prompt}, respectively.

\begin{table}[t]
  \centering
  \adjustbox{max width=\linewidth}{%
  \begin{tabular}{cc}
    \textbf{Prompt EN-IT} & \textbf{Prompt IT-EN} \\
    \midrule
    \makecell{Traduci dall'Inglese all'Italiano\\Text: I love you so much.\\Translation: Ti amo così tanto.} & \makecell{Translate from Italian to English\\Text: Ti amo così tanto.\\Translation: I love you so much.} \\
    \bottomrule
      \end{tabular}
  }
  \caption{Prompts used for machine translation task}
  \label{tab:FLoRes_prompt}
\end{table}

\begin{table}[t]
  \centering
  \adjustbox{max width=\linewidth}{%
  \begin{tabular}{c}
    \textbf{Italian Prompt}\\
    \midrule
    \makecell{Contesto: Il terremoto del Sichuan del 2008 o il terremoto\\ del Gran Sichuan, misurato a 8.0 Ms e 7.9 Mw,\\ e si è verificato alle 02:28:01 PM China ...\\Domanda: In quale anno si è verificato il terremoto nel Sichuan?\\Risposta: 2008} \\
    \bottomrule
      \end{tabular}
  }
  \caption{Used prompts for question answering task}
  \label{tab:SQuAD_prompt}
\end{table}

\section{English Results on Multi-choice benchmarks}
\label{app:ensligh_results}

In this section, we present a detailed analysis of the evaluation results on English benchmarks. \Cref{tab:zero_shot_en_mistral} reports the performance of \mistral{} on six multiple-choice benchmarks. From this table, we observe that \textit{SAVA} and \textit{FVT} achieve higher task-wise scores early in the adaptation process. A similar trend is evident for the \llama{} adapted models, as shown in \Cref{tab:zero_shot_en_llama}, where the \textit{SAVA} technique yields higher average scores than \textit{FVT}, at the beginning and at the end of training. For both models, per-task scores remain below the base model’s performance. However, incorporating a portion of English data during adaptation prevents catastrophic forgetting when transitioning towards the Italian language.

\begin{table*}[t]
  \centering
  \adjustbox{max width=0.9\linewidth}{%
  \begin{tabular}{lcccccc|c}
\hline
\textbf{Model} & \textbf{Hellaswag} & \textbf{MMLU} & \textbf{Arc Easy} & \textbf{PIQA} & \textbf{SCIQ} & \textbf{BOOLQ} & \textbf{AVG}\\\hline
\rowcolor{gray!20}Mistral-7B-v0.1 & $75.98_{\pm0.44}$ & $57.19_{\pm0.42}$ & $78.55_{\pm0.94}$ & $83.84_{\pm0.94}$ & $95.82_{\pm0.80}$ & $77.64_{\pm0.78}$ & 78.17 \\\hline
& \multicolumn{7}{c}{200 Training Steps} \\\hline
Random & $72.29_{\pm0.44}$ & $51.59_{\pm0.42}$ & $69.55_{\pm0.95}$ & $81.73_{\pm0.96}$ & $89.97_{\pm0.97}$ & $74.03_{\pm0.76}$ & 73.19 \\
FVT & $72.35_{\pm0.44}$ & $53.04_{\pm0.42}$ & $73.08_{\pm0.92}$ &$ 82.60_{\pm0.94}$ & $92.48_{\pm0.85}$ & $72.20_{\pm0.78}$ & \underline{74.29} \\
CLP & $72.59_{\pm0.44}$ & $52.02_{\pm0.42}$ & $70.16_{\pm0.94}$ & $81.55_{\pm0.96}$ & $89.66_{\pm0.98}$ & $72.81_{\pm0.77}$ & 73.13 \\
SAVA & $72.81_{\pm0.44}$ & $53.21_{\pm0.42}$ & $74.28_{\pm0.94}$ & $82.47_{\pm0.96}$ & $92.79_{\pm0.83}$ & $71.59_{\pm0.78}$ & \textbf{74.52} \\\hline
LAPT & $74.13_{\pm0.43}$ & $55.05_{\pm0.42}$ & $75.23_{\pm0.89}$ & $84.02_{\pm0.91}$ & $94.46_{\pm0.73}$ & $71.98_{\pm0.78}$ & 75.81 \\\hline
\rowcolor{cyan!20}& \multicolumn{7}{c}{2000 Training Steps} \\\hline
\rowcolor{cyan!20}Random & $72.18_{\pm0.44}$ & $52.11_{\pm0.42}$ &$ 73.6_{\pm0.91}$ & $82.72_{\pm0.94}$ & $93.21_{\pm0.81}$ & $75.77_{\pm0.74}$ & 74.93 \\
\rowcolor{cyan!20}FVT & $73.28_{\pm0.44}$ & $52.96_{\pm0.42}$ & $74.76_{\pm0.90}$ & $81.91_{\pm0.95}$ & $94.05_{\pm0.76}$ & $74.46_{\pm0.76}$ & \underline{75.23} \\
\rowcolor{cyan!20}CLP & $73.37_{\pm0.44}$ & $52.48_{\pm0.42}$ & $74.07_{\pm0.90}$ & $82.47_{\pm0.94}$ & $94.05_{\pm0.76}$ & $74.83_{\pm0.75}$ & 75.21 \\
\rowcolor{cyan!20}SAVA & $73.02_{\pm0.44}$ & $52.91_{\pm0.42}$ & $74.67_{\pm0.90}$ & $82.29_{\pm0.94}$  & $94.46_{\pm0.73}$ & $74.58_{\pm0.76}$ & \textbf{75.32} \\\hline
\rowcolor{cyan!20}LAPT & $74.26_{\pm0.43}$ & $51.18_{\pm0.42}$ &$ 73.9_{\pm0.91}$ & $83.65_{\pm0.92}$ & $94.67_{\pm0.72}$ & $74.22_{\pm0.76}$ & 75.31 \\\hline
  \end{tabular}
  }
  \caption{0-shot results over English benchmarks for \mistral{} adapted models.}
  \label{tab:zero_shot_en_mistral}
\end{table*}

\begin{table*}[t]
  \centering
  \adjustbox{max width=0.9\linewidth}{%
  \begin{tabular}{lcccccc|c}
\hline
\textbf{Model} & \textbf{Hellaswag} & \textbf{MMLU} & \textbf{Arc Easy} & \textbf{PIQA} & \textbf{SCIQ} & \textbf{BOOLQ} & \textbf{AVG} \\
\hline
\rowcolor{gray!20}LLaMa-3.1-8B & $74.21_{\pm0.43}$ & $62.19_{\pm0.42}$ & $77.60_{\pm0.47}$ & $83.03_{\pm0.93}$ & $93.94_{\pm0.77}$ & $80.42_{\pm0.69}$ & 78.56 \\\hline
& \multicolumn{7}{c}{300 Training Steps} \\\hline
FVT & $72.35_{\pm0.44}$ & $58.22_{\pm0.42}$ & $69.55_{\pm0.95}$ & $81.30_{\pm0.97}$ & $92.27_{\pm0.86}$ & $71.34_{\pm0.79}$ & 74.17 \\
SAVA & $72.72_{\pm0.44}$ & $58.19_{\pm0.42}$ & $70.75_{\pm0.94}$ & $81.79_{\pm0.96}$ & $92.90_{\pm0.83}$ & $71.28_{\pm0.79}$ & \textbf{74.60} \\\hline
LAPT & $74.35_{\pm0.43}$ & $61.74_{\pm0.41}$ & $76.14_{\pm0.88}$ & $83.21_{\pm0.93}$ & $94.05_{\pm0.76}$ & $76.69_{\pm0.73}$ & 77.69 \\\hline
\rowcolor{cyan!20}& \multicolumn{7}{c}{3000 Training Steps} \\\hline
\rowcolor{cyan!20}FVT & $73.02_{\pm0.44}$ & $57.85_{\pm0.42}$ & $72.13_{\pm0.93}$ & $82.04_{\pm1.15}$ & $92.90_{\pm0.83}$ & $72.53_{\pm0.78}$ & 75.07 \\
\rowcolor{cyan!20}SAVA & $72.86_{\pm0.44}$ & $57.94_{\pm0.42}$ & $72.78_{\pm0.92}$ & $81.79_{\pm0.96}$ & $93.31_{\pm0.80}$ & $73.30_{\pm0.77}$ & \textbf{75.33} \\\hline
\rowcolor{cyan!20}LAPT & $74.40_{\pm0.43}$ & $60.50_{\pm0.42}$ & $75.32_{\pm0.89}$ & $82.47_{\pm0.94}$ & $93.63_{\pm0.78}$ & $77.43_{\pm0.73}$ & 77.29 \\\hline
\end{tabular}
  }
  \caption{0-shot results over English benchmarks for \llama{} adapted models.}
  \label{tab:zero_shot_en_llama}
\end{table*}

\end{document}